\documentclass[conference]{IEEEtran}
\IEEEoverridecommandlockouts
\usepackage{cite}
\usepackage{amsmath,amssymb,amsfonts}
\usepackage{algorithmic}
\usepackage{textcomp}
\usepackage{xcolor}
\usepackage{booktabs}
\usepackage{graphicx}
\usepackage{subcaption}
\usepackage{hyperref}
\usepackage{xurl}

\usepackage{balance}

\def\BibTeX{{\rm B\kern-.05em{\sc i\kern-.025em b}\kern-.08em
    T\kern-.1667em\lower.7ex\hbox{E}\kern-.125emX}}
\begin{document}

\title{Why Convolution Still Matters: Evaluating Inductive Biases in Cryospheric Image Classification}

\author{
\IEEEauthorblockN{Chhaya Kulkarni}
\IEEEauthorblockA{
Department of Computer and Information Sciences\\
Towson University\\
ckulkarni@towson.edu
}
\and
\IEEEauthorblockN{Emam Hossain}
\IEEEauthorblockA{
Department of Information Systems\\
University of Maryland, Baltimore County\\
emamh1@umbc.edu
}
}

\maketitle

\begin{abstract}


Recent advances in attention-based deep learning have motivated their adoption for remote sensing image classification; however, their benefits for cryospheric imagery, where surface states are dominated by fine-grained textures and class imbalance, remain unclear. In this work, we revisit a benchmark Greenland Ice Sheet image dataset, previously shown to favor convolutional neural networks (CNNs), to examine whether modern attention-based and hybrid architectures improve class-wise reliability. We conduct a controlled comparison between a classical CNN (AlexNet), a modern CNN (ConvNeXt-Tiny), a pure attention-based model (Swin-Tiny), and a hybrid convolution–attention model (CoAtNet-0) under identical training and evaluation protocols. Results show that AlexNet achieves the highest accuracy and the strongest balanced performance as measured by macro-averaged F1, while ConvNeXt-Tiny exhibits highest macro-averaged AUC, indicating strong class separability but less consistent final decision quality. Class-wise analysis reveals that hybrid architectures improve recall for rare and structurally distinct surface classes, whereas convolutional models remain more reliable for texture-dominated categories. These findings highlight the importance of aligning architectural inductive bias with cryospheric data characteristics and suggest that increased model complexity does not necessarily translate to improved reliability for ice-sheet surface classification.


\end{abstract}


\section{Introduction}

Remote sensing of the cryosphere is essential for monitoring surface processes such as snow accumulation, melt, refreezing, and ice dynamics. Automated image classification methods are widely used to support large-scale analysis of these processes, particularly in regions where manual interpretation is impractical. Due to the prevalence of fine-grained textures, subtle intensity variations, and spatial heterogeneity, convolutional neural networks (CNNs) have been widely adopted for cryospheric image classification tasks \cite{dunmire2021contrasting, liu2019review, kortum2024, kucik2023ai4seaice}.

Accurate classification of cryospheric surface types is not an end in itself but a critical intermediate step for downstream scientific analysis and operational monitoring.
Distinguishing among surface classes associated with melt, refreezing, accumulation, and structural ice features supports improved interpretation of surface energy balance, meltwater routing, and seasonal surface evolution in ice-covered regions \cite{kimball2021comparing, li2025assessing}.
Errors in rare or transitional surface classes can disproportionately affect estimates of melt extent and surface condition trends, particularly in regions undergoing rapid climate change.
Consequently, evaluating class-wise performance and error patterns is essential for assessing the reliability of classification models used in cryospheric applications.

Several prior studies have reported strong performance using relatively simple CNN architectures, including AlexNet \cite{krizhevsky2012imagenet}, on benchmark cryospheric image datasets \cite{devon_dunmire_2021_4813833}. These findings suggest that convolutional inductive bias particularly sensitivity to local texture and spatial patterns aligns well with the visual characteristics of cryospheric imagery.

In parallel, attention-based deep learning models and Vision Transformers have gained popularity across computer vision and remote sensing due to their ability to model long-range dependencies and global context \cite{dosovitskiy2020image, liu2021swin}. Recent work has applied such architectures to remote sensing image classification and land cover mapping, reporting competitive performance on large-scale datasets with strong global structure \cite{li2025assessing,le2024onboard,roy_2023}.
However, their effectiveness for texture-dominated cryospheric imagery remains unclear. Hybrid architectures that combine convolutional operations with attention mechanisms, such as CoAtNet-0 \cite{dai2021coatnet}, aim to integrate local feature extraction with global contextual reasoning. Although effective on general vision benchmarks, their suitability for cryospheric image classification remains less established.


Despite growing interest in attention-based and hybrid models, comparative studies that evaluate classical CNNs, modern CNNs, pure attention-based models, and hybrid architectures under identical training and evaluation protocols are limited in the cryospheric domain. Moreover, existing studies often emphasize aggregate accuracy while overlooking the convergence behavior, class imbalance, and class-wise error characteristics that are critical for reliable cryospheric interpretation.


\begin{figure*}[!h]
    \centering
    \includegraphics[width=\textwidth]{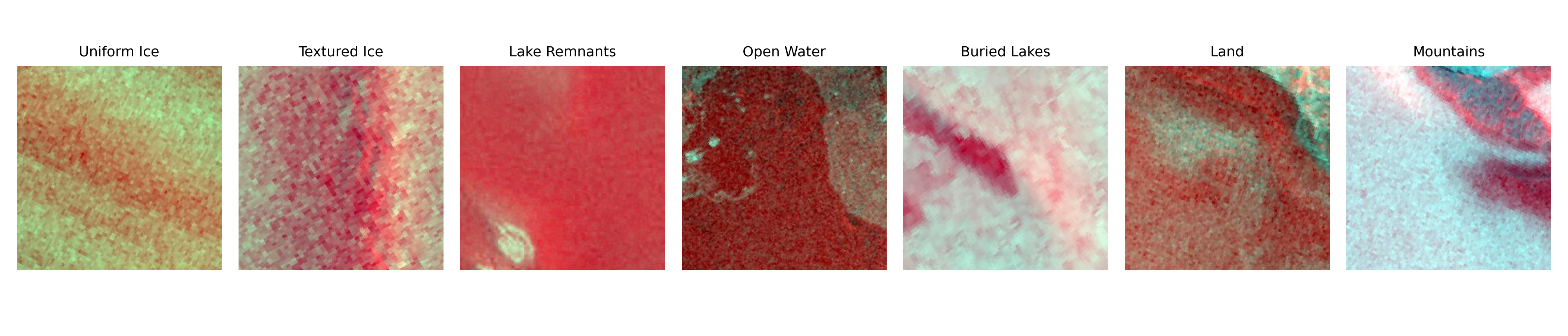}
    \caption{Representative image tiles for the seven cryospheric surface classes used in this study, following the class definitions of Dunmire et al.~\cite{dunmire2021contrasting}. The examples highlight variability in surface texture, spatial structure, and melt-related signatures across classes, as well as the visual similarity between several ice surface categories.}
    \label{fig:class_examples}
\end{figure*}

In this work, we revisit a benchmark cryospheric image dataset~\cite{devon_dunmire_2021_4813833} previously shown to favor AlexNet and conduct a controlled architectural comparison between a classical CNN (AlexNet~\cite{krizhevsky2012imagenet,zhou2019remote}), a modern CNN (ConvNeXt-Tiny~\cite{liu2022convnext,liu2025efficient}), a pure attention-based model (Swin-Tiny\cite{jiang2023remote}), and a hybrid convolution–attention model (CoAtNet-0~\cite{dai2021coatnet, aldahoul2023space}). All models are trained using identical data splits, optimization settings, and early stopping criteria, and are evaluated using class-balanced metrics. Rather than proposing a new architecture or claiming state-of-the-art performance, our objective is to clarify how architectural inductive bias influences convergence behavior, class-wise discrimination, and balanced performance in cryospheric image classification. 


\section{Related Work}


Initial applications of deep learning in cryospheric image classification primarily utilized convolutional neural networks, exhibiting significant advancements compared to traditional machine learning methods. These studies emphasized the importance of local spatial features and texture sensitivity, with AlexNet often serving as a strong baseline due to its simplicity and effectiveness \cite{dunmire2021contrasting, liu2019review, kumar2025analyzing}. Subsequent work in remote sensing expanded CNN-based approaches through deeper architectures such as ResNet and DenseNet, reporting improved representation learning while retaining convolutional inductive bias \cite{adegun2023review,fayaz2024land}. However, many of these studies evaluated a single architecture in isolation, limiting insight into architectural trade-offs.

More recently, attention-based models and Vision Transformers have been applied to remote sensing imagery, particularly for scene classification and land cover mapping tasks involving large datasets and globally structured patterns \cite{kumari2023recent, he2022swin}. These models exploit self-attention to capture long-range dependencies, offering a fundamentally different inductive bias compared to CNNs. Their performance on texture-dominated cryospheric imagery remains underexplored.


Hybrid convolution–attention architectures, such as CoAtNet-0, seek to integrate convolutional locality with global contextual reasoning and have demonstrated strong results on general vision benchmarks \cite{dai2021coatnet}. While recent frameworks are beginning to extract joint spectral-spatial features via hybrid designs \cite{arshad2024}, controlled comparisons with classical and modern CNNs remain scarce in the cryospheric domain. Existing work lacks a unified evaluation of CNN, attention-based, and hybrid architectures for cryospheric image classification, which this study addresses through a controlled comparison.



\section{Methodology}

\subsection{Dataset and Preprocessing}

This study uses the cryospheric image dataset introduced by Dunmire et al.~\cite{devon_dunmire_2021_4813833}, which was developed to support automated classification of surface and near-surface melt features across the Greenland Ice Sheet. The dataset consists of labeled image tiles derived from Sentinel-1 synthetic aperture radar (SAR) and Sentinel-2 optical observations, combined to capture complementary information on surface texture, scattering behavior, and melt-related structure.

Each image tile is assigned to one of seven surface classes representing distinct cryospheric surface states and geomorphological features: \textit{uniform ice and snow} (1,277 tiles), \textit{textured ice and snow} (921 tiles), \textit{surface lake remnants} (1,322 tiles), \textit{open ocean/water} (99 tiles), \textit{buried lakes} (1,400 tiles), \textit{land} (542 tiles), and \textit{mountains surrounded by snow} (227 tiles). The final experimental dataset used in this study contains 5,788 image tiles across the seven classes.




\begin{figure*}[!h]
    \centering
    \includegraphics[width=\textwidth]{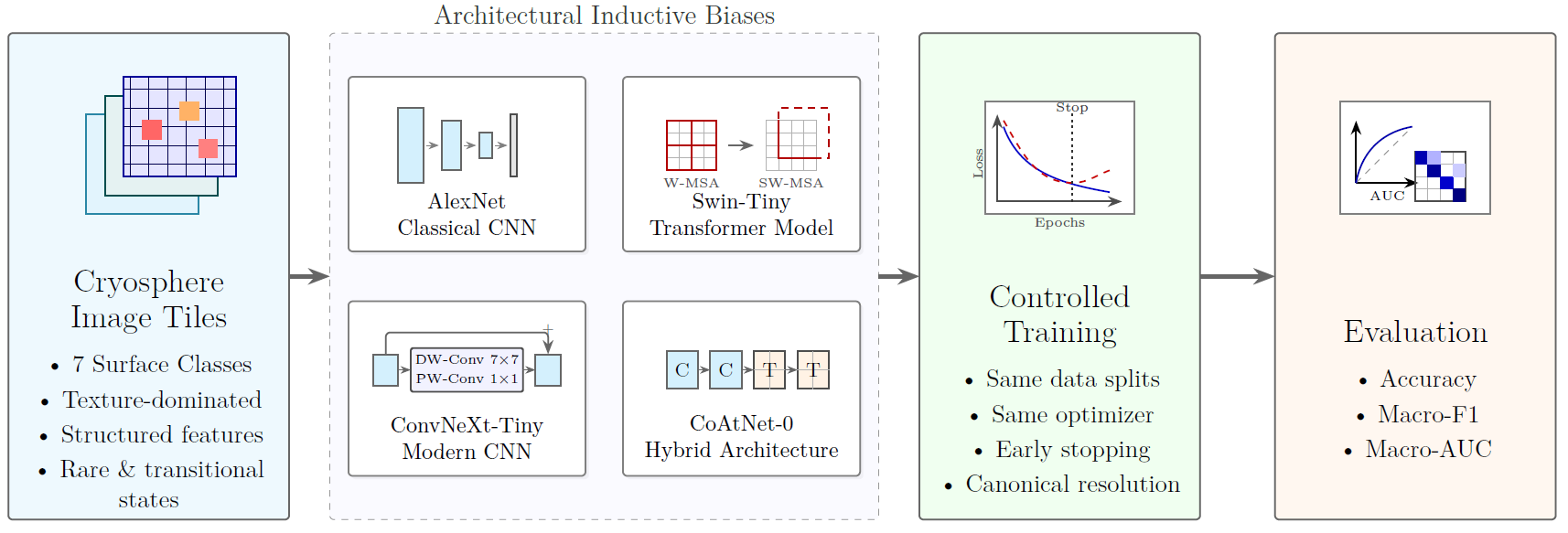}
    \caption{Overview of the experimental framework for controlled comparison of convolutional, attention-based, and hybrid architectures for cryospheric image classification. All models are trained under identical data splits and optimization settings and evaluated using class-balanced performance metrics.}
    \label{fig:methodology}
\end{figure*}

All image tiles are normalized using ImageNet mean and standard deviation statistics to match pretrained model initialization. Convolutional architectures (AlexNet and ConvNeXt-Tiny) are trained using $256 \times 256$ pixel inputs, consistent with prior cryospheric CNN studies, while attention-based and hybrid architectures (Swin-Tiny and CoAtNet-0) are trained at their canonical $224 \times 224$ input resolution to respect architectural design and pretraining conventions. Apart from random horizontal flipping, no additional data augmentation is applied to isolate architectural effects.


Representative examples of the seven surface classes are shown in Fig.~\ref{fig:class_examples}, illustrating the range of textural and structural variability present in the dataset and the visual similarity between several ice surface categories that make this classification task challenging.

\subsection {Model Architectures, Training, and Evaluation}

Four architectures are evaluated to represent distinct architectural inductive biases: AlexNet as a classical convolutional baseline, ConvNeXt-Tiny as a modern CNN, Swin-Tiny as a pure attention-based model, and CoAtNet-0 as a hybrid convolution–attention architecture. Here, inductive bias refers to architectural assumptions that shape how spatial patterns are learned, with convolution emphasizing local texture and attention emphasizing global context. All models are trained using categorical cross-entropy loss and the \textit{Adam} optimizer with a fixed learning rate of $5 \times 10^{-4}$. \textit{Adam} is used to maintain a common optimization framework across CNN, transformer-based, and hybrid architectures. Early stopping based on validation loss is employed to ensure that all models are evaluated at convergence rather than at an arbitrary epoch count, enabling fair comparison across architectures with differing convergence dynamics. Batch size is adjusted per architecture based on memory requirements (32 for convolutional models, 16 for Swin-Tiny, and 8 for CoAtNet-0), following standard practice. Fig.~\ref{fig:methodology} presents an overview of the overall methodology used for controlled architectural comparison in cryospheric image classification.

The dataset is split into training, validation, and held-out test sets using a 70\%/15\%/15\% split, with 4,012 tiles for training, 888 for validation, and 888 for testing. Performance is evaluated on the held-out test set using three metrics: accuracy, Macro-F1, and Macro-AUC, with Macro-F1 emphasized due to class imbalance. Confusion matrices are analyzed to examine class-specific error patterns relevant to cryospheric interpretation. Experiment restricting all models to 100 training epochs using early stopping is conducted to illustrate differences in convergence behavior. To facilitate reproducibility, the source code and experimental configurations for this study are publicly available on GitHub\footnote{\url{https://github.com/ChhayaKulkarni/cryospheric-image-classification}}.

\section{Results and Discussion}


Table~\ref{tab:performance} reports the performance on the test set for all four evaluated architectures using accuracy, macro-averaged F1 score (Macro-F1), and macro-averaged area under the receiver operating characteristic curve (Macro-AUC), with Macro-F1 serving as the primary metric to account for class imbalance.

\begin{table}[!h]
\centering
\caption{Overall classification performance on the test set.}
\label{tab:performance}
\begin{tabular}{lccc}
\toprule
\textbf{Model} & \textbf{Accuracy} & \textbf{Macro-F1} & \textbf{Macro-AUC} \\
\midrule
AlexNet        & 0.912 & 0.901 & 0.986 \\
ConvNeXt-Tiny  & 0.900 & 0.884 & 0.991 \\
Swin-Tiny      & 0.848 & 0.826 & 0.984 \\
CoAtNet-0      & 0.870 & 0.853 & 0.988 \\
\bottomrule
\end{tabular}
\end{table}

\begin{figure*}[!t]
    \centering
    \begin{subfigure}[b]{0.24\textwidth}
        \centering
        \includegraphics[width=\linewidth]{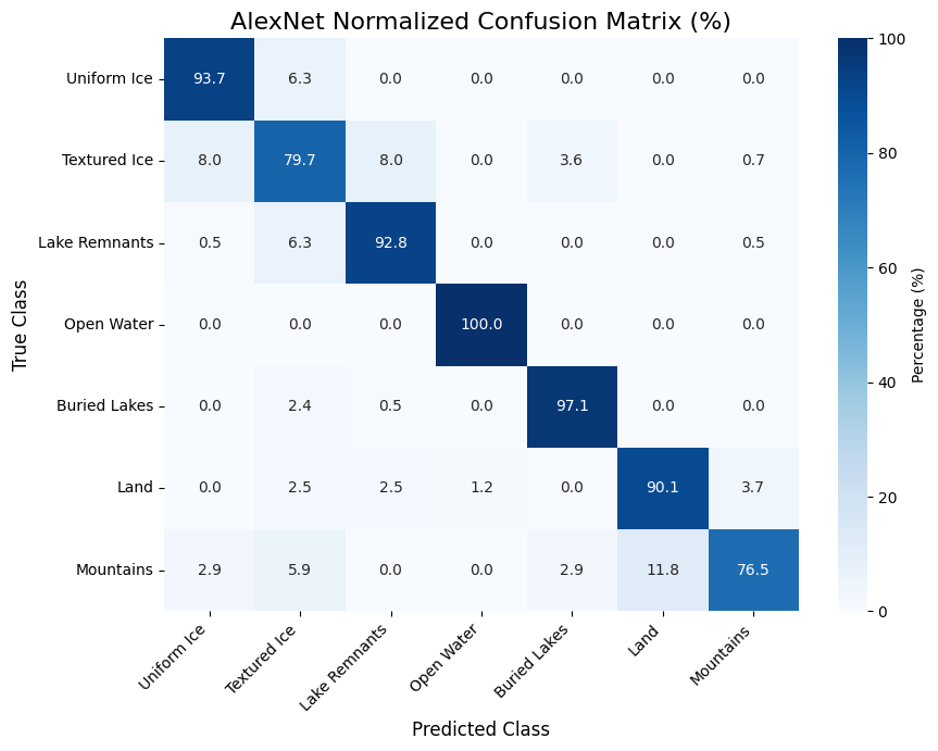}
        \caption{AlexNet}
    \end{subfigure}
    \hfill
    \begin{subfigure}[b]{0.24\textwidth}
        \centering
        \includegraphics[width=\linewidth]{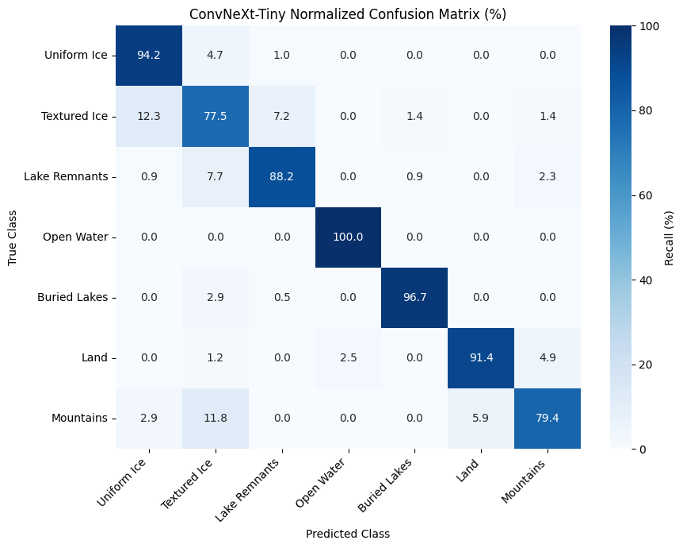}
        \caption{ConvNeXt-Tiny}
    \end{subfigure}
    \hfill
    \begin{subfigure}[b]{0.24\textwidth}
        \centering
        \includegraphics[width=\linewidth]{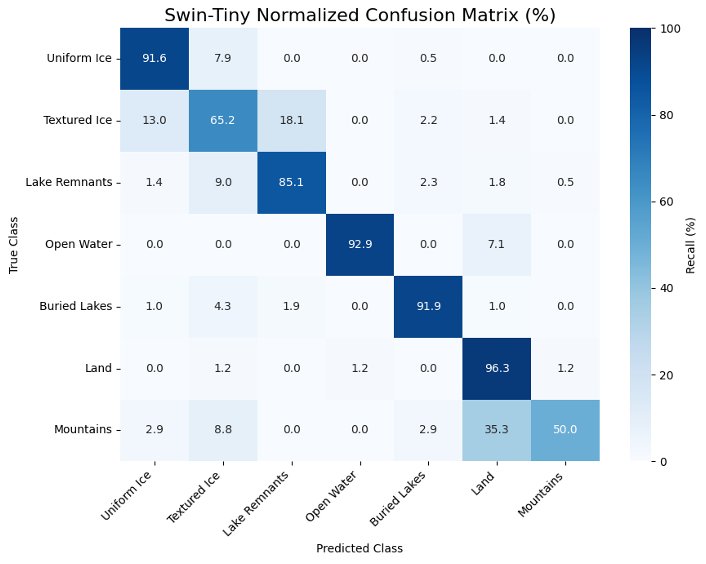}
        \caption{Swin-Tiny}
    \end{subfigure}
    \hfill
    \begin{subfigure}[b]{0.24\textwidth}
        \centering
        \includegraphics[width=\linewidth]{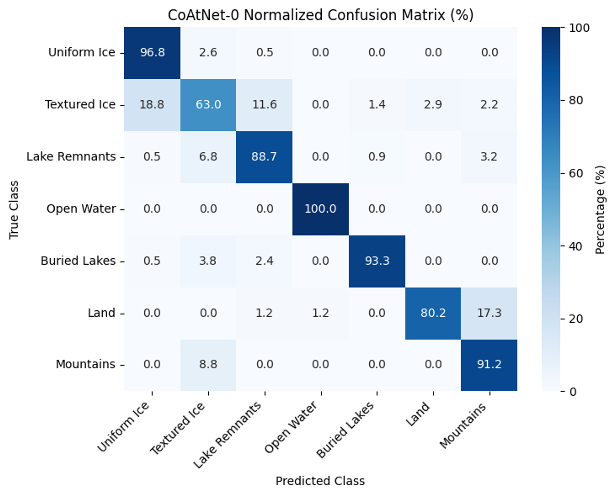}
        \caption{CoAtNet-0}
    \end{subfigure}

    \caption{Recall-normalized confusion matrices (\%) for the four evaluated architectures. Rows correspond to true classes and columns to predicted classes; diagonal entries indicate per-class recall.}
    \label{fig:confusion_matrix}
\end{figure*}

\textbf{Overall performance trends:} Across all models, convolutional architectures achieve the strongest balanced classification performance. AlexNet attains the highest Macro-F1 score (0.901), closely followed by ConvNeXt-Tiny (0.884), indicating that convolutional inductive bias remains highly effective for cryospheric surface classification. Despite its relative architectural simplicity, AlexNet generalizes well, suggesting strong alignment between local texture sensitivity and the dominant visual characteristics of the dataset. ConvNeXt-Tiny achieves the highest Macro-AUC (0.991), reflecting superior class separability; however, this advantage does not consistently translate into higher Macro-F1, highlighting a distinction between representation separability and final decision quality under class imbalance.

\textbf{Attention-based and hybrid architectures:}  Transformer-based architectures exhibit lowest performance across all the performance metrics. Swin-Tiny records the lowest accuracy (0.848), Macro-F1 score (0.826) and Macro-AUC (0.984), particularly struggling on texture-dominated and visually ambiguous surface categories. While self-attention enables modeling of long-range spatial dependencies, these results suggest that global attention alone is insufficient for reliable discrimination of fine-grained cryospheric textures when training data are moderate in size.

The hybrid CoAtNet-0 architecture consistently improves upon Swin-Tiny across all reported metrics, demonstrating the benefit of integrating convolutional locality with attention-based global reasoning. CoAtNet-0 achieves higher accuracy (0.870), Macro-F1 (0.853) and Macro-AUC (0.988) than Swin-Tiny, indicating a more balanced trade-off between class separability and decision quality. Nevertheless, hybrid designs do not surpass CNN-based architectures in overall performance.


\textbf{Class-wise error analysis:} Figure~\ref{fig:confusion_matrix} presents recall-normalized confusion matrices for all four architectures, enabling direct inspection of class-specific behavior. CNN-based models (AlexNet and ConvNeXt-Tiny) exhibit cleaner diagonals and fewer diffuse confusions, particularly for texture-dominated classes such as \textit{textured ice} and \textit{lake remnants}, as well as for the rare and scientifically important \textit{buried lakes} class. This behavior reflects the effectiveness of convolutional feature extraction for capturing local spatial variability characteristic of melt-related surface features. 

Transformer-based models show improved recall for globally structured terrain classes, such as \textit{land} and \textit{mountains}. CoAtNet-0 achieves the most balanced per-class recall across surface types, while Swin-Tiny exhibits substantially reduced recall for \textit{mountains}, with frequent misclassification as \textit{land}. This failure mode indicates limitations of windowed self-attention in capturing large-scale topographic structure without convolutional support. \textit{Open water} shows near-perfect recall due to strong visual separability.

\textbf{Training dynamics and architectural implications:} All models are trained for up to 100 epochs with early stopping based on validation loss. At convergence, the CNN-based AlexNet achieves the highest test accuracy (0.912) and Macro-F1 (0.901), while ConvNext-Tiny achieves the highest Macro-AUC (0.991). In contrast, Swin-Tiny performs the lowest across the reported metrics. These results suggest that convolutional and hybrid convolution-attention architectures are better suited than the pure attention-based model for this cryospheric image classification task.


Overall, these results indicate that architectural inductive bias plays a more decisive role than model complexity for cryospheric image classification. Convolutional locality remains critical for reliable discrimination of texture-dominated and rare surface types, while attention mechanisms enhance performance on globally structured terrain. Hybrid architectures partially mitigate the limitations of pure attention-based models but do not surpass CNN-based approaches when balanced performance across all classes is required. These findings show that architectural inductive bias shapes class-balanced performance across cryospheric surface types, with convolution supporting texture discrimination and attention contributing global context.

\section{Conclusion and Future Work}

This study conducted a controlled comparison of convolutional, attention-based, and hybrid deep learning architectures for cryospheric image classification using a benchmark dataset previously shown to favor CNN-based models. Under identical training and evaluation protocols, classical and modern CNN architectures achieved the strongest balanced performance, as measured by Macro-F1. Attention-based and hybrid models demonstrated high class separability and improved recall for certain rare and structured classes but did not consistently outperform CNNs in overall classification quality.

These results indicate that architectural inductive bias plays a key role in cryospheric image classification. While attention mechanisms enhance representation learning, convolutional locality remains critical for discriminating texture-dominated surface classes. Hybrid architectures improve upon pure transformer models by integrating convolutional features but remain sensitive to ambiguity in mid-frequency classes. 

This study focuses on compact (“tiny”) variants of attention-based and hybrid architectures to ensure fair comparison under similar computational budgets and to mitigate overfitting on a moderate-sized dataset. While larger-capacity models may further improve performance, increasing model scale does not fundamentally alter architectural inductive bias. Future work will therefore investigate larger-scale and multi-scale attention models, as well as physics-informed and multi-modal extensions, to further assess the role of global context in cryospheric image classification.

\section*{Acknowledgment}

This work is supported by iHARP: NSF HDR Institute for Harnessing Data and Model Revolution in the Polar Regions (Award\# 2118285). The views expressed in this work do not necessarily reflect the policies of the NSF, and endorsement by the Federal Government should not be inferred.

\balance
\bibliographystyle{IEEEtran}
\bibliography{refs}

\end{document}